\documentclass[10pt,twocolumn,letterpaper]{article}

\usepackage{iccv}
\usepackage{times}
\usepackage{epsfig}
\usepackage{graphicx}
\usepackage{amsmath}
\usepackage{amssymb}

\usepackage{enumitem}
\usepackage{booktabs}
\usepackage{nicefrac}
\newcommand{\todo}[1]{}
\renewcommand{\todo}[1]{{\color{red} TODO: {#1}}}
\usepackage{float}

\newcommand{\convlayer}{\includegraphics{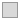} }
\newcommand{\uppelayer}{\includegraphics{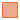} }
\newcommand{\uppelayerr}{\includegraphics{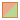} }
\newcommand{\downlayer}{\includegraphics{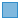} }
\newcommand{\downlayerr}{\includegraphics{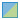} }
\newcommand{\flatlayer}{\includegraphics{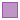} }

\usepackage[pagebackref=true,breaklinks=true,letterpaper=true,colorlinks,bookmarks=false]{hyperref}

\iccvfinalcopy 

\def\iccvPaperID{0237} 

\ificcvfinal\fi
\begin{document}

\title{Learning Blind Motion Deblurring}

\author{Patrick Wieschollek$^{1,2}$ 
  \and
  Michael Hirsch$^2$ 
  \and
  Bernhard Sch\"olkopf$^2$
  \and
  Hendrik P.A. Lensch$^1$ \\ [3mm]
  $^1$ University of T{\"u}bingen\\
    $^2$ Max Planck Institute for Intelligent Systems, T{\"u}bingen\\
}

\maketitle

\begin{abstract}
As handheld video cameras are now commonplace and available in every smartphone, images and videos can be recorded almost everywhere at anytime.
However, taking a quick shot frequently yields a blurry result due to unwanted camera shake during recording or moving objects in the scene. 
Removing these artifacts from the blurry recordings is a highly ill-posed problem as neither the sharp image nor the motion blur kernel is known. 
Propagating information between multiple consecutive blurry observations can help restore the desired sharp image or video. 
In this work, we propose an efficient approach to produce a significant amount of realistic training data and introduce a novel recurrent network architecture to deblur frames taking temporal information into account, which can efficiently handle arbitrary spatial and temporal input sizes. 

\end{abstract}

\section{Introduction}
Videos captured by handheld devices usually contain motion blur artifacts caused by a combination of camera shake (ego-motion) and dynamic scene content (object motion). With a fixed exposure time any movement during recording causes the sensor to observe an averaged signal from different points in the scene.
A reconstruction of the sharp frame from a blurry observation 
is a highly ill-posed problem, denoted as blind or non-blind deconvolution depending on whether camera-shake information is known or not.

In video and image burst deblurring the reconstruction process for a single frame can make use of additional data from neighboring frames. However, the problem is still challenging as each frame might encounter a different camera shake and the frames might not be aligned.

For deconvolution of a static scene neural networks have been successfully applied using single frame~\cite{chakrabarti,schuler2013machine,Schuler_PAMI15} and multi-frame deblurring \cite{burstdeblurring,sparseprior,fba}.

All recent network architectures for multi-frame and video deblurring \cite{burstdeblurring, deepvideo, NorooziCF17, chakrabarti} require the input to match a fixed temporal \textit{and} spatial size. Handling arbitrary spatial dimensions is theoretically possible by fully convolutional networks as done in \cite{deepvideo}, but they rely on a sliding window approach during inference due to limited memory on the GPU. For these approaches, the reconstruction of one frame is not possible by aggregating the information of longer sequences than the network was trained for.

In contrast, our approach is a deblurring system that can deal with arbitrary lengths of sequences while featuring a fully convolutional network that can process full resolution video frames at once. Due to its small memory footprint it removes the need for sliding window approaches during inference, thus drastically accelerating the deblurring process. 
For processing arbitrary sequences we rely on a recurrent scheme. While convolutional LSTMs \cite{convLSTM} offer a straightforward way to replace spatial convolutions in conventional architectures by recurrent units, we found them challenging and slow to train. Besides vanishing gradients effects they require a bag of tricks like carefully tuned gradient clipping parameters and a special variant of batch normalization. 
In order to circumvent these problems, we introduce a new recurrent encoder-decoder network.
In the network we incorporate spatial residual connections and introduce novel temporal feature transfer between subsequent iterations.

Besides the network architecture, we further create a novel training set for video deblurring as the success of data-driven approaches heavily depends on the amount and quality of available realistic training examples.
As acquiring realistic ground-truth data is time-consuming, we successfully generate synthetic training data with literally no acquisition cost and demonstrate improved results and run time on various benchmark sets as for example demonstrated in Figure~\ref{fig:teaser}.

\section{Related Work}

The problem of image deblurring can be formulated as a non-blind or a blind deconvolution version, depending on whether information about the blur is available or not. Blind image deblurring (BD) is quite common in real-world applications and has seen considerable progress in the last decade.
A comprehensive review is provided in the recent overview article by Wang and Tao \cite{wang2014recent}. Traditional state-of-the-art methods such as Sun \etal~\cite{sun2013edge} or Michaeli and Irani \cite{michaeli2014blind} use carefully chosen patch-based priors for sharp image prediction. Data-driven methods based on neural networks have demonstrated success in non-blind restoration tasks \cite{schuler2013machine,convnetdeblur,rosenbaum2015return} as well as for the more challenging task of BD where the blur kernel is unknown \cite{Schuler_PAMI15,sun2015learning,chakrabarti,hradivs2015convolutional,svoboda2016cnn}. Removing the blur from moving objects has been recently addressed in~\cite{NorooziCF17}.

To alleviate the ill-posedness of the problem \cite{hasinoff2009time}, one might take multiple observations into account. Hereby, observations of a static scene, each of which is differently blurred, serve as inputs \cite{sparseprior,chen2008robust,cai2009blind,vsroubek2012robust,zhu2012deconvolving,hirsch2010efficient}. 
To incorporate video properties such as temporal consistency the methods of \cite{zhang2014multi,zhang2015intra,kim2016dynamic,ito2014blurburst} use powerful and flexible generative models to explicitly estimate the unknown blur along with predicting the latent sharp image. However, this comes at the price of higher computation cost, which typically requires tens of minutes for the restoration process.

To accomplish faster processing times Delbracio and Sapiro \cite{delbracio2015hand} have presented a clever way to average a sequence of input frames based on Lucky Imaging methods. They propose to compute a weighted combination of all aligned input frames in the Fourier domain which favors stable Fourier coefficients in the burst containing sharp information.
This yields much faster processing times and removes the requirement to compute the blur kernel explicitly. 

Quite recently, Wieschollek \etal in \cite{burstdeblurring} introduce an end-to-end trainable neural network architecture for multi-frame deblurring. Their approach directly computes a sharp image by processing the input burst in a patch-wise fashion, yielding state-of-the-art results. It has been shown, that this even enables treating spatially varying blur. The related task of deblurring of videos has been approached by Su \etal~\cite{deepvideo}. Their approach uses the U-Net architecture \cite{unet} with skip connection to directly regress the sharp image from an input burst. Their fully convolutional neural network learns an average of multiple inputs with reasonable performance. Unfortunately, both learning methods \cite{burstdeblurring,deepvideo} require to fix the temporal input size at training time and they are limited to a patch-based inference by the network layout \cite{burstdeblurring} and memory constraints \cite{deepvideo}.

\section{Method}

\paragraph{Overview.} In our approach a fully-convolutional neural network deblurs a frame $\mathcal{I}$ using information from previous frames $\mathcal{I}_{-1}, \mathcal{I}_{-2},\ldots$ in an iterative, recurrent fashion.
Incorporating a previous (blurry) observation improves the current prediction for $\mathcal{I}$ step by step. We will refer to these steps as \textit{deblur steps}. Hence, the complete \textit{recurrent deblur network} (RDN) consists of several deblur blocks (DB).
We use weight-sharing between these to reduce the total amount of used parameters and introduce novel temporal skip connections between these deblur blocks to propagate latent features between the individual temporal steps. To effectively update the network parameters we unroll these steps during training.
At inference time, the inputs can have arbitrary spatial dimensions as long as the processing of a minimum of two frames fits on the GPU. Moreover, the recurrent structure allows us to include an arbitrary number of frames helping to improve the output with each iteration. Hence, there is no need for padding burst sequences to match the network architecture as e.g.\ in \cite{burstdeblurring}.

\subsection{Generating realistic ground-truth data}

Training a neural network to predict a sharp frame of a blurry input requires realistic training data featuring these two aligned versions for each video frame: a blurry version serving as the input and an associated sharp version serving as ground-truth. Obtaining this data is challenging as any recorded sequence might suffer from the described blur effects itself. 
Recent work \cite{deepvideo,NorooziCF17} have built a training data set by recording videos captured at 240fps with a GoPro Hero camera to minimize the blur in the ground-truth. Frames from these high-fps videos are then processed and averaged to produce plausible motion blur synthetically. 
While they made significant effort to capture a broad range of different situations, this process is limited in the number of recorded samples, in the variety of scenes and in the used recording devices. For fast moving objects artifacts are likely to arise due to the finite framerate.  
We also tested this method for generating training data with a GoPro Hero camera but found it hard to produce a large enough dataset of sharp ground-truth videos of high quality. Rather than acquiring training data manually we propose to acquire and filter data from online media.

\begin{figure}[tb]
  \centering
  \includegraphics[width=.48\textwidth]{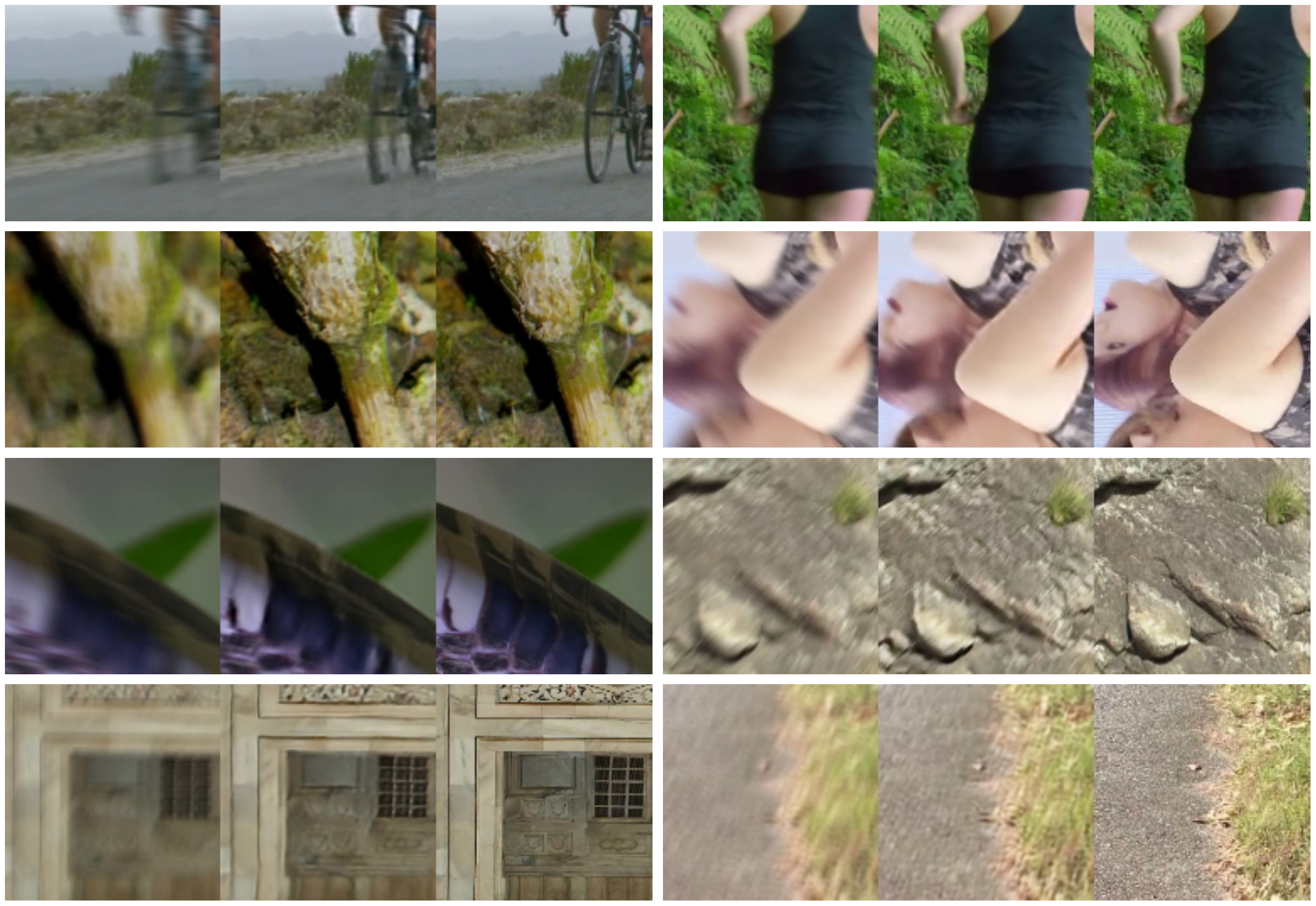}
  \caption{Snapshot of the training process. Each triplet shows the input with synthetic blur (left), the current network prediction (middle) and the associated ground-truth (right). All images are best viewed at higher resolution in the electronic version. }
  \label{fig:training_examples}
\end{figure}

\paragraph{Training data.}
As people love to share and rate multimedia content, each year millions of video clips are uploaded to online platforms like YouTube. The video content ranges from short clips to professional videos of up to 8k resolution. From this source, we have collected videos with 4k-8k resolution and a frame rate of 60fps or 30fps. The video content ranges from movie trailers, sports events, advertisements to videos on everyday life. To remove compression artifacts and to obtain slightly sharper ground-truth we resized all collected videos by factor $\nicefrac{1}{4}$ respectively $\nicefrac{1}{8}$, finally obtaining full-HD resolution.

Consider such a video with frames $(f_t)_{t=1,2,\ldots, T}$. For each frame pair $(f_t, f_{t+1})$ at time $t$ we compute $n$ additional \textit{synthetical} subframes between the original frames $\mathbf{f_t}, \mathbf{f_{t+1}}$ resulting in a high frame rate video
$$(\ldots, f_{t-1}^{(n-1)}, f_{t-1}^{(n)}, \mathbf{f_t}, f_t^{(1)},f_t^{(2)},\ldots, f_t^{(n-1)}, f_t^{(n)},\mathbf{f_{t+1}}).$$ 
All subframes are computed by blending between the neighboring original frames $f_t$ and $f_{t+1}$ warping both frames using the optical flow in both directions $w_{f_t\to f_{t+1}}$ and $w_{f_{t+1}\to f_{t}}$. Given both flow fields, we can generate an arbitrary number of subframes. For practical purposes, we set $n=40$, thus implying an effective framerate of more than 1000fps without suffering from low signal-to-noise ratio (SNR) due to short exposure times.

We want to stress that only parts of videos with reasonably sharp frames serve as ground-truth. For those the estimation of optical flow to approximate motion blur is possible and sufficient. The sub-frames are averaged to generate a plausible blurry version
\begin{align}
\label{eq:averageframes}
    b_t = \frac{1}{1+2L}\left( \mathbf{f_t} + \sum_{\ell=1}^L 
    f_{t-1}^{(n-\ell)} + f_{t}^{(\ell)}\right)
\end{align}
for each sharp frame $f_t$. We use a mix of 20 and 40 for L to create different levels of motion blur.
The entire computation can be done offline on a GPU.
For all video parts that passed our sharpness test (5.43 hours in total) we produce a ground-truth video and blurry version both at 30 fps in full-HD. 
Besides the unlimited amount of training data another major advantage of this method is that it incorporates different capturing devices naturally. Further, the massive amount of available video content allows us to tweak all thresholds and parameters in a conservative way to reject video parts of bad quality (too dark, too static) without affecting the effective size of the training data. 
Though the recovered optical flow is not perfect we observed an acceptable quality of the synthetically motion blurred dataset. To add variety to the training data we crop random parts from the frames and resize them to 128$\times$128px. Figure~\ref{fig:training_examples} shows a few random examples from our training dataset.

\subsection{Handling the time dimension}
The typical input shape required by CNNs in computer vision tasks is $[B,H,W,C]$ -- batch size, height, width and number of channels. However, processing series of images includes a new dimension: time. To apply spatial convolution layers the additional dimension has to be ``merged'' either into 
the channel $[B,H,W,C\cdot T]$ or batch dimension $[B\cdot T,H,W,C]$. 
Methods like \cite{deepvideo,burstdeblurring} stack the time along the channel dimension rendering all information across the entire burst available without further modification. This comes at the price of removing  information about the temporal order. Further, the number of input frames needs to be fixed before training, which limits their application. Longer sequences could only be processed with workarounds like padding and sliding window processing. On the other hand, merging the time-dimension into the batch dimension would give flexibility at processing different length of sequences. 
But the processing of each frame is then entirely decoupled from its adjacent frames -- no information is propagated. Architectures using convLSTM \cite{convLSTM} or convGRU cells \cite{grucell} are designed to naturally handle time series but they would require several tricks \cite{batchnorm_rnn,dropconnect} during training. We tried several architectures based on these recurrent cells but found them hard to train and observed hardly any improvement even after two days of training.

\begin{figure*}[h!]
  \centering
  \includegraphics[width=.9\textwidth]{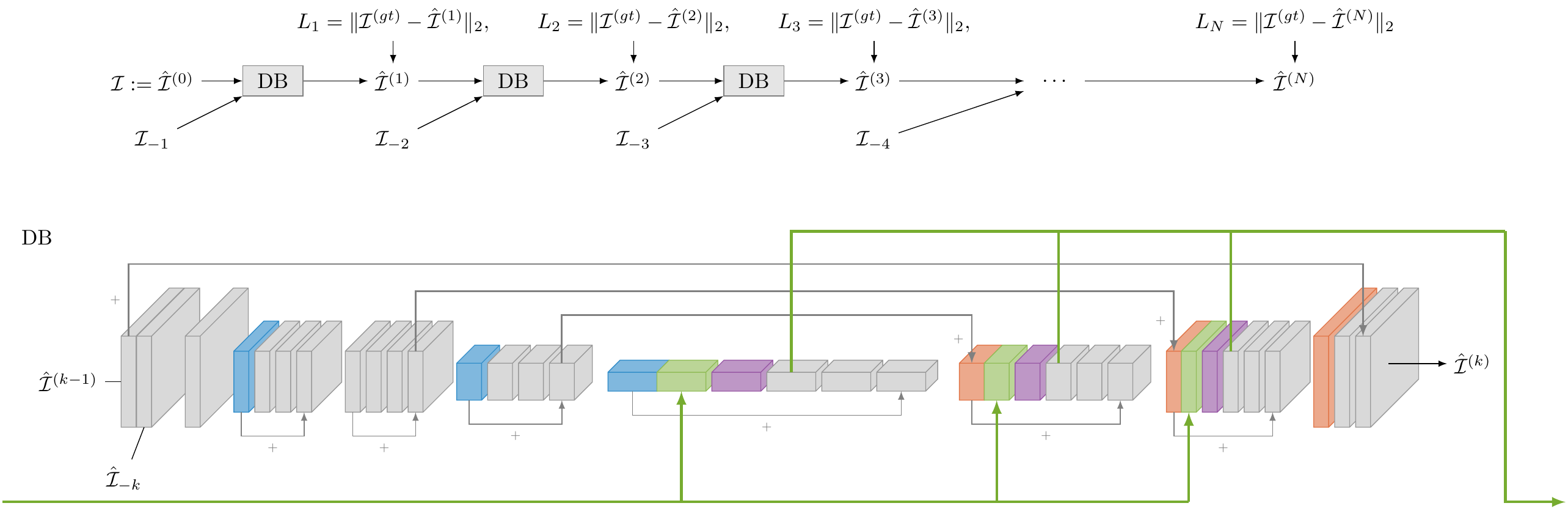}
  \caption{Given the current deblurred version of $\mathcal{I}$ each deblur block DB produces a sharper version of $\mathcal{I}$ using information contributed by another observation $\mathcal{I}_{-k}$. The deblur block follows the design of an encoder-decoder network with several residual blocks with skip-connections. To share learned features between various observations, we propagate some previous features into the current DB (green).}
  \label{fig:network}
\end{figure*}

\subsection{Network Architecture}

Instead of including recurrent layers, we propose to formulate the entire network as a recurrent application of deblur blocks and successively process pairs of inputs (target frame and additional observation), which gives us the flexibility to handle arbitrary sequence lengths and enables information fusion inside the network. 

Consider a single deblur step with the current prediction $\mathcal{I}$ of shape $[H,W,C]$ and blurry observation $\mathcal{I}_{-k}$.
Inspired by the work of Ronneberger \etal \cite{unet} and the recent success of residual connections \cite{resnet} we use an encoder-decoder architecture in each deblur block, see Figure~\ref{fig:network}. Hereby, the network only consists of convolution and transpose-convolution layers with batchnorm \cite{batchnorm15}. We applied the ReLU activation to the input of the convolution layers $C_{\cdot,\cdot}$ as proposed in \cite{resnet}.

\begin{table}[h]
  \caption{Network Specification. Outputs of layers marked with * are concatenated with features from previous deblur blocks except in the first step. This doubles the channel size of the output. The blending layers $B_{\cdot, \cdot}$ are only used after the first deblur step.}
  \label{tab:network}
  \centering

  \begin{tabular}{llcl}
  \toprule
   layer               & filter size            & stride            & output shape    \\
  \midrule
   \convlayer $A_{0,1}$           & $3\times 3 \times 64 $ & 2                 & $\nicefrac{H}{1} \times \nicefrac{H}{1} \times 64 $\\
  \midrule
   \downlayer $C_{1,1}$           & $3\times 3 \times 64 $ & 2                 & $\nicefrac{H}{2} \times \nicefrac{H}{2} \times 64 $\\
   \convlayer $C_{1,2} \!- \!C_{1,4}$ & $3\times 3 \times 64 $ & 1                 & $\nicefrac{H}{2} \times \nicefrac{H}{2} \times 64 $\\
  \midrule
   \convlayer $C_{2,1} \!- \!C_{2,4}$ & $3\times 3 \times 64 $ & 1                 & $\nicefrac{H}{2} \times \nicefrac{H}{2} \times 64 $\\
  \midrule
   \downlayer $C_{3,1}$           & $3\times 3 \times 128$ & 2                 & $\nicefrac{H}{4} \times \nicefrac{H}{4} \times 128 $\\
   \convlayer $C_{3,1} \!- \!C_{3,4}$ & $3\times 3 \times 128$ & 1                 & $\nicefrac{H}{4} \times \nicefrac{H}{4} \times 128 $\\
  \midrule
   \downlayerr $C_{4,1}$           & $3\times 3 \times 256$ & 2                 & $\nicefrac{H}{8} \times \nicefrac{H}{8} \times 256 $*\\
   \flatlayer $B_{4,2}$           & $1\times 1 \times 256$ & 1                 & $\nicefrac{H}{8} \times \nicefrac{H}{8} \times 256 $\\
   \convlayer $C_{4,3} \!- \!C_{4,5}$ & $3\times 3 \times 256$ & 1                 & $\nicefrac{H}{8} \times \nicefrac{H}{8} \times 256 $\\
  \midrule
   \uppelayerr $C_{5,1}$           & $4\times 4 \times 128$ & $\nicefrac{1}{2}$ & $\nicefrac{H}{4} \times \nicefrac{H}{4} \times 128 $*\\
   \flatlayer $B_{5,2}$           & $1\times 1 \times 128$ & 1                 & $\nicefrac{H}{4} \times \nicefrac{H}{4} \times 128 $\\
   \convlayer $C_{5,3} \!- \!C_{5,5}$ & $3\times 3 \times 128$ & 1                 & $\nicefrac{H}{4} \times \nicefrac{H}{4} \times 128 $\\
  \midrule
   \uppelayerr $C_{6,1}$           & $4\times 4 \times 64 $ & $\nicefrac{1}{2}$ & $\nicefrac{H}{2} \times \nicefrac{H}{2} \times 64 $*\\
   \flatlayer $B_{6,2}$           & $1\times 1 \times 64 $ & 1                 & $\nicefrac{H}{2} \times \nicefrac{H}{2} \times 64 $\\
   \convlayer $C_{6,3} \!- \!C_{6,5}$ & $3\times 3 \times 64 $ & 1                 & $\nicefrac{H}{2} \times \nicefrac{H}{2} \times 64 $\\
  \midrule
   \uppelayer $C_{7,1}$           & $4\times 4 \times 64 $ & $\nicefrac{1}{2}$ & $\nicefrac{H}{1} \times \nicefrac{H}{1} \times 64 $\\
   \convlayer $C_{7,2}$           & $4\times 4 \times 6 $  & 1                 & $\nicefrac{H}{1} \times \nicefrac{H}{1} \times 6 $\\
   \convlayer $\mathcal{I}^{(k)}$ & $3\times 3 \times 3 $  & 1                 & $\nicefrac{H}{1} \times \nicefrac{H}{1} \times 3 $\\
  \bottomrule
  \end{tabular}
\end{table}

The first trainable convolution layer expands the 6-channel input (two $128\times 128$px RGB images during training) into 64 channels.
In the encoder part, each residual block consists of a down-sampling convolution layer (\downlayer\!\!) followed by three convolution layers (\flatlayer\!\!). The down-sampling layer halves the spatial dimension with stride 2 and doubles the effective number of channels $[H,W,C]\to [H/2,W/2,C\cdot2]$. 
During the decoding step, the transposed-convolution layer (\uppelayer\!\!) inverts the effect of the downsampling $[H,W,C]\to [2\cdot H,2\cdot W,C/2]$. We use a filter size of $3\times3$ / $4\times4$ for all convolution/transposed-convolution layers. In the beginning an additional residual block without downsampling accounts for resolving larger blur by providing a larger receptive field. 

To speed up the training process, we add skip-connections between the encoding and decoding part. Hereby, we add the extracted features from the encoder to the related decoder part. This enables the network to learn a residual between the blurry input and the sharp ground-truth rather than ultimately generating a sharp image from scratch. Hence, the network is fully-convolutional and therefore allows for arbitrary input sizes. Please refer to Table~\ref{tab:network} for more details.

\paragraph{Skip connections as temporal links.}
We also propose to propagate latent features between subsequent deblur blocks over time. For this, we concatenate specific layer activations from a previous iteration with some from the current deblur block. These skip connections are illustrated as green lines in Figure~\ref{fig:network}. Further, to reduce the channel dimension to match the required input shape for the next layer, we use a $1\times 1$ convolution layer, denoted as blending layer $B_{\cdot, \cdot}$. This way the network can learn a weighted sum by blending between the current features and propagated features from the previous iteration. This effectively halves the channel dimension.
One advantage of such a construction is that we can disable these skip connections in the first deblur block and only apply these in subsequent iterations. Further, they can be applied to a pre-trained model without temporal skip connections.

\paragraph{Training details.}  Aligning inputs using homography matrices or estimated optical flow information can be error-prone and slows down the reconstruction preventing time-critical applications. Therefore, we trained the network directly on a sequence of unaligned frames featuring large camera shakes. To further challenge the network we add artificial camera shake to each blurry frame from synthetic PSF kernels on-the-fly.
These PSF kernels of sizes $7\times 7, 11\times 11, 15\times 15$ are generated by a Gaussian process simulating camera shake. To account for the effect of vanishing gradients, we force the output $\mathcal{I}^{(k)}$ of each deblur block to match the sharp ground-truth $\mathcal{I}^{(gt)}$ in the corresponding loss term $L_k$ (see Figure~\ref{fig:network}). 

We use ADAM \cite{adam} for minimzing the total loss ${L=\sum_{k=1}^4 L_k}$ for sequences of 5 inputs. We leave the optimizer's default parameters ($\beta_1= 0.9$, $\beta_2 = 0.999$) unchanged and use 5e-3 as the initial learning rate.

\section{Experiments}
We evaluate the performance of our proposed method in several experiments on challenging real-world examples. In addition, a comprehensive comparison to recent methods is given using the implementation provided by the respective authors. During inference we pass a pair of frames with resolution 720p into a deblur block iteratively. Each iteration takes approximately 0.57 seconds on an NVIDIA Titan X. For any larger frame sizes, we tile the input frames. The network was trained exclusively on our synthetically blurred dataset featuring both motion blur and camera shake. All provided results in the section are based on benchmark sets from previous methods. Our \textit{recurrent deblur network} (RDN) generalizes to different kinds of unseen videos and recording devices. Please note, we include the full-resolution images and frames from videos in the supplementary material.

\subsection{Burst Deblurring}

\begin{figure*}[!h]
  \centering
  \includegraphics[width=\textwidth]{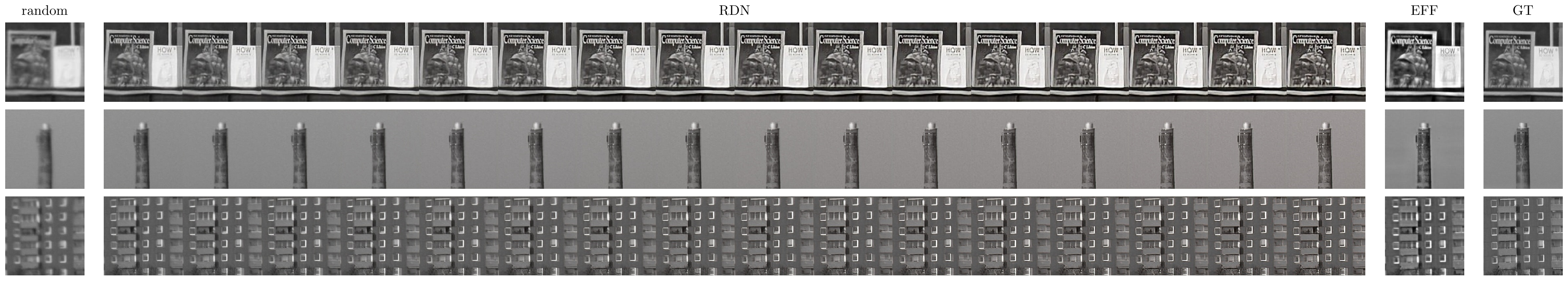}
  \caption{Recovery from image bursts with spatially varying blur. Reconstructions of a single image using 2 to 17 input frames are shown. Although the network has only seen training sequences of length 5, due to its recurrent structure it can handle longer sequences and further improve the prediction. However, too many input frames might introduce oversharpening. A random shot from the input is given on the left and the ground-truth on the right. We also compare against the EFF reconstructions from Hirsch \etal~\cite{hirsch2010efficient}, which is dedicated to this task. }
  \label{fig:eff_comparison}
\end{figure*}
\begin{figure*}[tb]
  \centering
  \includegraphics[width=\textwidth]{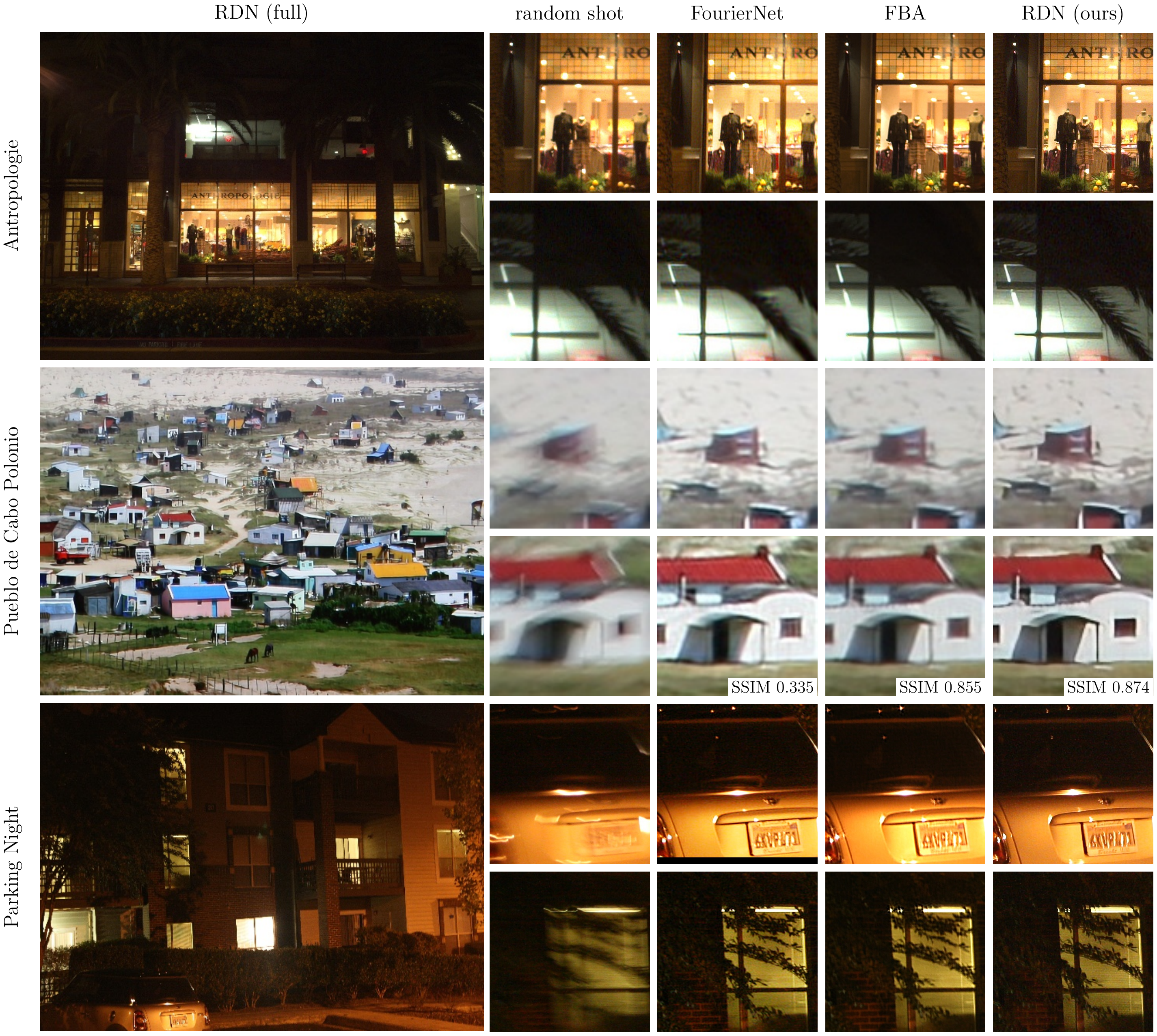}
  \caption{Comparison to state-of-the-art multi-frame blind deconvolution algorithms FourierNet~\cite{burstdeblurring}, FBA~\cite{delbracio2015hand} and ours (RDN) on real-world data for static scenes of low-light environments. RDN recovers significantly more detail.}
  \label{fig:burst_deblurring}
\end{figure*}

\begin{figure}[]
  \centering
  \includegraphics[width=.48\textwidth]{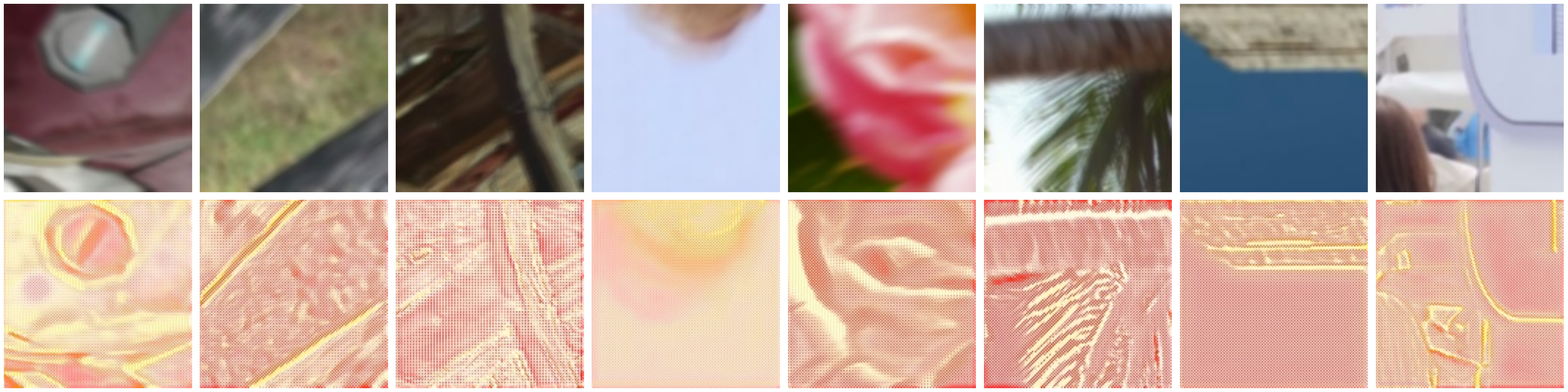}
  \caption{Visualization of features propagated along the temporal skip connections. The additional channels are projected to 2D and encoded in Hue colorspace. Apparently, the network learned to mark regions which might benefit from deblurring.}
  \label{fig:blur_field}
\end{figure}
\begin{figure}[]
  \centering
  \includegraphics[width=.48\textwidth]{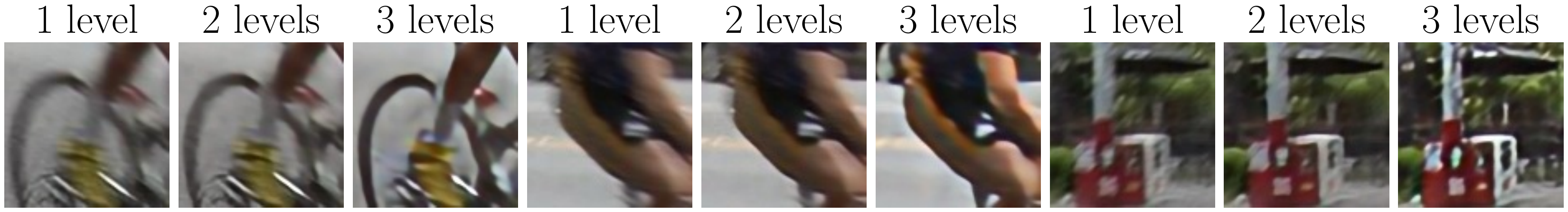}
  \caption{Multi-scale input for large motion blur. We show the deblurred results with tradition single-scale input (1 level) or extending the input sequence with upscaled version of the deblurred results at half respectively quarter resolution.}
  \label{fig:multiscale}
\end{figure}

\begin{figure}[H]
  \centering
  \includegraphics[width=.48\textwidth]{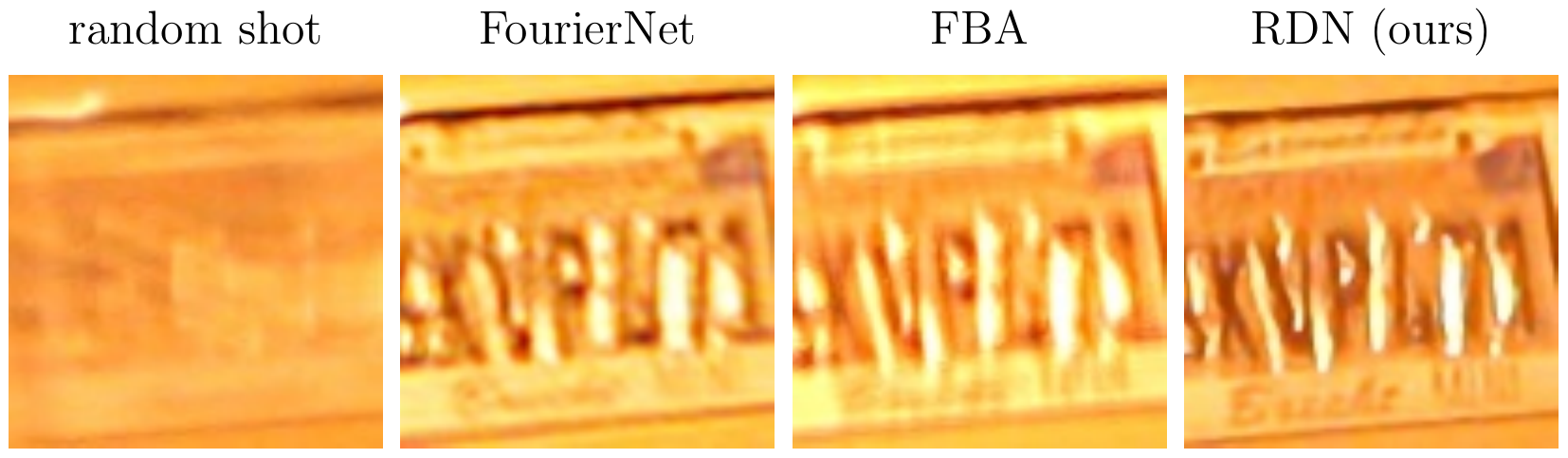}
  \caption{In contrast to previous state-of-the-art methods, our recurrent approach is able to even recover the subtle writing on the bottom of this number plate. It further reflects the original color tones from the random blurry shot.}
  \label{fig:burst_licenseplate}
\end{figure}

\begin{figure*}[]
  \centering
  \includegraphics[width=\textwidth]{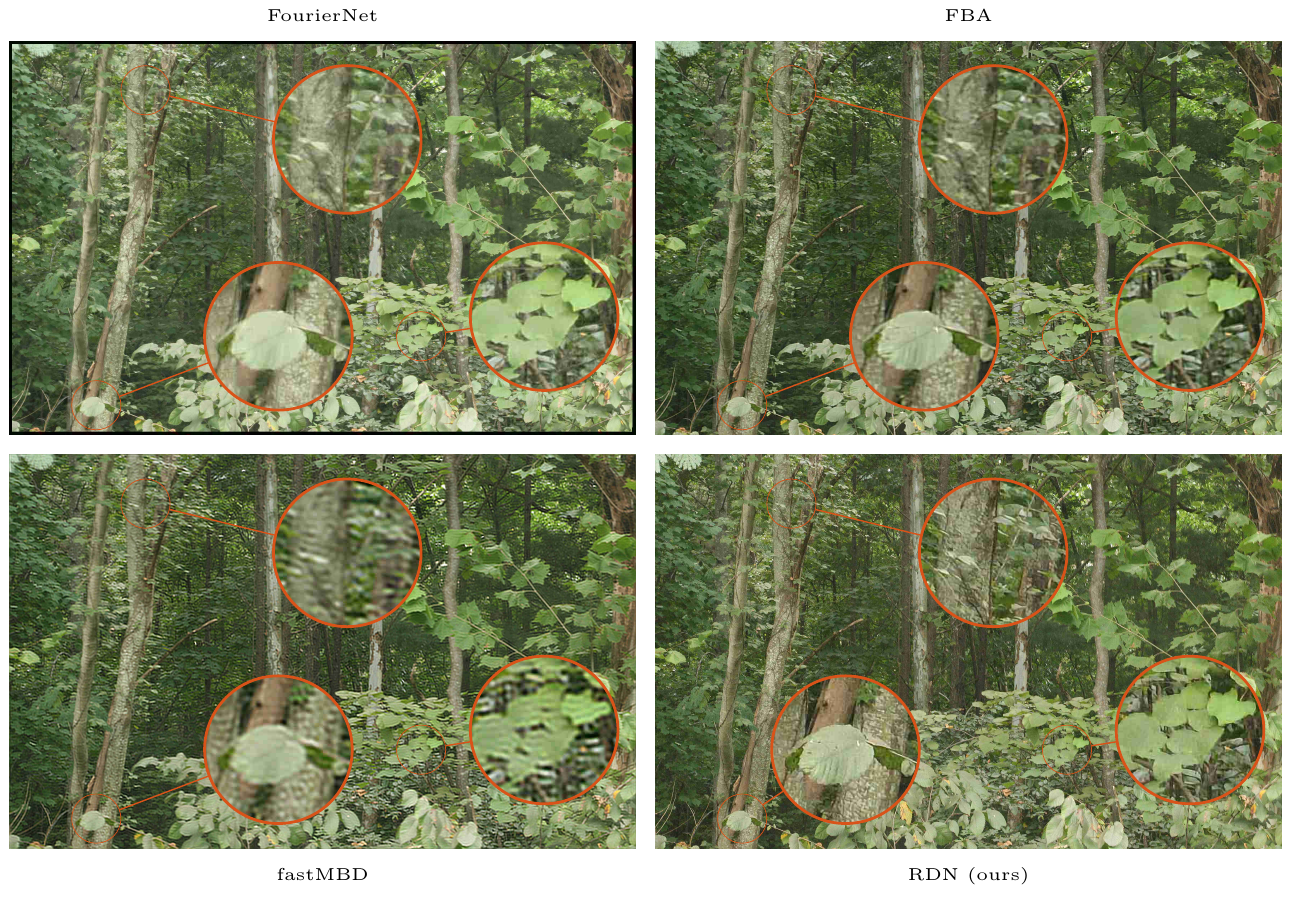}
  \caption{Performance of multi-frame blind deconvolution algorithms FourierNet~\cite{burstdeblurring}, FBA~\cite{delbracio2015hand} and ours (RDN) on a forest scene. The structure of the leaves and the bark of the trees is significantly sharper.}
  \label{fig:accv_comparison}
\end{figure*}

In burst deblurring the task is to restore a sharp frame from an entire sequence of aligned images. The sequence is usually taken by a single camera and only suffers from stationary blur caused by ego-motion. In our data-driven approach, we process each observation which finally produces significantly better results than previous proposed methods, \eg consider the scene provided in Figure~\ref{fig:burst_deblurring}. Notably, ours is the first, which is able to restore the lettering below the license plate in Figure~\ref{fig:burst_licenseplate}. Also for the wood scene in Figure~\ref{fig:accv_comparison} sharper results are produced.

Further, our network is applied to input images featuring spatially varying blur, which is quite common in real-world examples due to imperfect lenses or turbulences. Figure~\ref{fig:eff_comparison} shows a comparison to the Efficient Filter Flow framework (EFF) \cite{hirsch2010efficient} which is explicitly designed to model this kind of blur. 
The results demonstrate two features of our approach: it is able to generalize to this kind of blur -- no patch-wise processing as in \cite{burstdeblurring} was necessary -- and due to its recurrent nature it can deal and exploit almost arbitrary many frames for deblurring one image. However, we observe in the top row of Figure~\ref{fig:eff_comparison} that after adding more than 10 input images local contrast might saturate, potentially resulting also in a small color shift. A workaround might be some color-transfer method \cite{burstdeblurring}.

\subsection{Video Deblurring}
\begin{figure}[H]
  \centering
  \includegraphics[width=.48\textwidth]{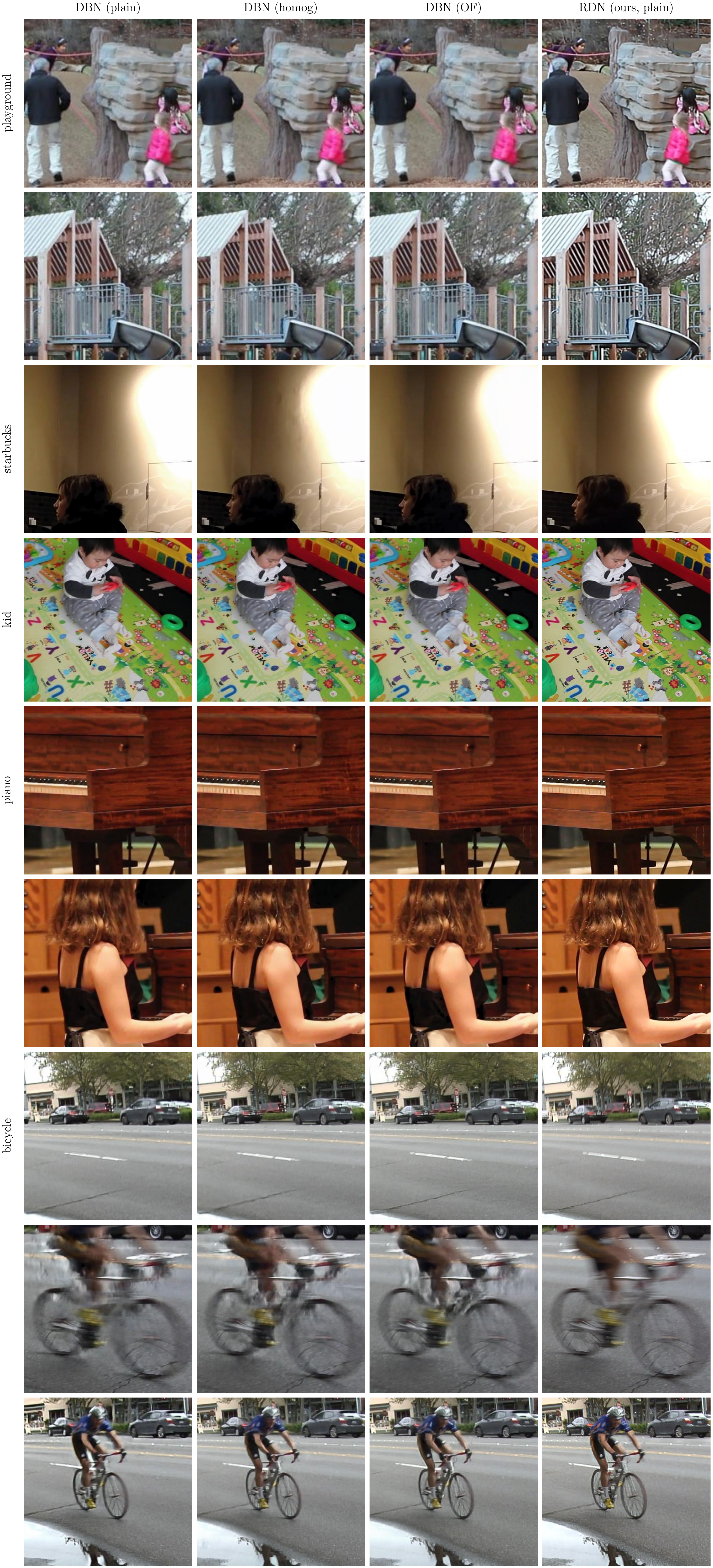}
  \caption{Video deblurring results. Applied to successive video frames with camera shake and motion blur our network produces favorable results compared to the approach by Su \etal~\cite{deepvideo} (DBN). While their performance depends on the type of offline preprocessing to align the individual frames (plain, homography, optical flow) our method operates directly on the unaltered input frames. As the preprocessing might introduce artifacts such as blur, smearing etc.\ their network learned to partially correct those, which sometimes fails. The strong motion blur of the cyclist is not correctly deblurred by any existing method.}
  \label{fig:video_deblurring}
\end{figure}

In contrast to the previous task, videos are usually degraded by additional blur caused by object motion. Moreover, any deblurring approach has to solve the underlying frame alignment problem. Such an alignment step can be done offline, \eg using a homography matrix or by estimating optical flow fields to warp the frames to the reference frame. While this kind of preprocessing delivers an easier task to the network, it might introduce artifacts which the network later has to account for. The approach of Su \etal~\cite{deepvideo} (DBN) extensively use preprocessing for alignment and directly train their networks to solve both tasks: deblurring and removing artifacts. Our approach does not require any preprocessing and is therefore faster while producing comparable or better results. Figure~\ref{fig:video_deblurring} shows a comparison between the DBN in \cite{deepvideo} and our network directly applied to the input. Significant improvement in sharpness by our method can be observed on the trousers, the hair of the woman or the hand of the baby, just to highlight a few. Artifacts due to the alignment procedure in the approach by Su \etal are visible on the lit wall in the Starbucks scene, for the cyclist in the second last row and in the piano scene, where the white keys are distorted. While ours is competitive when removing small motion, their optical flow based methods produces slightly sharper results when the camera motion is severe as seen on the road markings in the in the ``bicycle'' scene. Due to the limited capacity of the trained networks neither their nor our approach is fully capable of recovering the strong motion blur of very fast motion.

\paragraph{Using multi-scale input.} While our network has been trained on sequences of constant spatial resolution only, we experimented with feeding multi-scale input to recover strong object motion. In particular, we deblurred the entire input sequence at different levels $n=1,2,3$ with $\nicefrac{1}{2^{n-1}}$ resolution and then up-scaled the predicted result to obtain an additional new input frame for the sequence at the higher scale. While it partly helped to deal with larger motion blur which is not covered in the training data, the upsampling can produced artifacts which the network was not trained for. Figure~\ref{fig:multiscale} shows such results. Although the bike became significant sharper, the static parts of the scene rendered a ``comic style'' appearance. Directly training such a multi-scale network seems to be an interesting research direction.

\paragraph{Time-structure.} 
Our network architecture consists of an ``anti-causal'' structure deblurring one frame by considering the original previous frames in a sequence-to-one mapping. 
$\hat{\mathcal{I}} = DB(DB(\mathcal{I}, \mathcal{I}_{-1}), \ldots)$. 
We experimented with several sequence-to-sequence mapping approaches producing a sharp frame in an online way $\hat{\mathcal{I}}_{t} = DB(\mathcal{I}_{t}, \hat{\mathcal{I}}_{t-1})$. We noticed no learning benefit which might be caused by the limited capability of propagating temporal information.

\paragraph{Identifying valuable temporal information.} 
One novel feature of our designed network architecture are the temporal skip connections (Figure~\ref{fig:network} in green) acting as information links between subsequent deblur blocks. As we do not add constraints to these links, we essentially allow the network to propagate whatever feature information seems to be beneficial for the next deblur block. To illustrate these temporal information, we visualized the respective layer activation in Figure~\ref{fig:blur_field}. The illustration suggests that the network uses this opportunity to propagate image locations which might profit from further deblurring (yellowish parts).

\section{Conclusion}
We presented a novel recurrent network architecture -- recurrent deblurring network (RDN) -- for the efficient removal of blur caused by both ego and object motion from a sequence of unaligned blurry frames. Our proposed model enables fast processing of image sequences of arbitrary length and size. We introduce the concept of temporal skip connections between consecutive deblur blocks which allow for efficient information propagation across several time steps. Our proposed network iteratively improves the sharpness of a target frame given various blurry observations. 

Furthermore, we presented a novel method for the efficient generation of a vast number of blurry/sharp video sequence pairs, which is required to train learning based methods like the one we described. Using bidirectional optical flow between consecutive frames, our method creates synthetically intermediate frames to fake high-speed video recordings. By averaging multiple consecutive frames we can emulate longer exposure times and thus motion blur in a realistic way. Hence, making use of the abundance of high-quality videos available on YouTube, we illustrated the generation process of an arbitrary amount of training data.


\paragraph{Acknowledgement} This work was supported by the German Research Foundation (DFG): SFB  1233, Robust Vision: Inference Principles and Neural Mechanisms, TP XX. 



\newpage
{\small
\bibliographystyle{ieee}
\bibliography{egbib}
}

\end{document}